\documentclass{article}

\usepackage{neurips_2023}

\usepackage[utf8]{inputenc} 
\usepackage[T1]{fontenc}    
\usepackage{hyperref}       
\usepackage{url}            
\usepackage{booktabs}       
\usepackage{amsfonts}       
\usepackage{nicefrac}       
\usepackage{microtype}      

\usepackage{amsmath}
\usepackage{xcolor}         
\usepackage{array, booktabs, multirow}
\usepackage{graphicx}
\usepackage{authblk}
\newcommand*{\affaddr}[1]{#1} 
\newcommand*{\affmark}[1][*]{\textsuperscript{#1}}
\newcommand*{\email}[1]{\texttt{#1}}
\usepackage{caption}
\usepackage{subcaption}

\newcommand{\FIDLAr}{{\sc FIDLAr}}

\title{The Power of Explainability in Forecast-Informed Deep Learning Models for Flood Mitigation}

\author{%
\textbf{Jimeng Shi}\affmark[1], 
\textbf{Vitalii Stebliankin}\affmark[1], 
\textbf{Giri Narasimhan}\affmark[1] \\
\affaddr{\affmark[1]Florida International University} \\
\email{\{jshi008,vsteb002,giri\}@fiu.edu} \\
}

\begin{document}

\maketitle

\begin{abstract}
Floods can cause horrific harm to life and property.
However, they can be mitigated or even avoided by the effective use of hydraulic structures such as dams, gates, and pumps. 
By pre-releasing water via these structures in advance of extreme weather events, water levels are sufficiently lowered to prevent floods.
In this work, we propose \FIDLAr, a \underline{\textbf{F}}orecast \underline{\textbf{I}}nformed \underline{\textbf{D}}eep \underline{\textbf{L}}earning \underline{\textbf{Ar}}chitecture, achieving flood management in watersheds with hydraulic structures in an optimal manner by balancing out flood mitigation and unnecessary wastage of water via pre-releases. 
We perform experiments with \FIDLAr\ using data from the South Florida Water Management District, which manages a coastal area that is highly prone to frequent storms and floods.
Results show that \FIDLAr\ performs better than the current state-of-the-art with several orders of magnitude speedup and with provably better pre-release schedules.
The dramatic speedups make it possible for \FIDLAr\ to be used for real-time flood management.
The main contribution of this paper is the effective use of tools for model explainability, allowing us to understand the contribution of the various environmental factors towards its decisions.
\end{abstract}

\section{Introduction}
Floods can result in catastrophic loss of life \cite{jonkman2008loss}, huge socio-economic impact \cite{wu2021new}, property damage \cite{brody2007rising}, and environmental devastation \cite{yin2023flash}.
While flood risks may be on the rise in both frequency and scale because of global climate change\cite{wing2022inequitable, hirabayashi2013global}, the resulting sea-level rise in coastal areas \cite{sadler2020exploring} amplify the threats posed by floods. Therefore, improved and real-time flood management is of utmost significance.
Managing the \emph{control schedules} of hydraulic structures, such as dams, gates, pumps, and reservoirs \cite{kerkez2016smarter} can make controlled flood mitigation possible \cite{bowes2021flood}. 

Currently, decades of human experience on specific river systems have resulted in rule-based methods \cite{bowes2021flood, sadler2019leveraging} to decide control schedules.
However, the lack of sufficient experience to deal with extreme events leaves us vulnerable to catastrophic floods.
Additionally, the schedules may not generalize to complex river systems \cite{schwanenberg2015open}. 
%
Soft optimization methods and other physics-based models, which are currently used, are prohibitively slow \cite{leon2020matlab, sadler2019leveraging, vermuyten2018combining, chen2016dimension, leon2014dynamic}.

Machine learning (ML) has emerged as a powerful approach for this domain \cite{willard2023time}.
Although ML-based methods have been used for flood prediction \cite{mosavi2018flood, shi2023deep}, flood detection \cite{tanim2022flood, shahabi2020flood}, susceptibility assessment \cite{saha2021flood, islam2021flood}, and post-flood management \cite{munawar2019after}, they have not been used for flood mitigation. 
In this paper, we address this gap by applying well-engineered ML methods to the flood mitigation problem. 
\FIDLAr, a \underline{\textbf{F}}orecast \underline{\textbf{I}}nformed \underline{\textbf{D}}eep \underline{\textbf{L}}earning \underline{\textbf{Ar}}chitecture, is trained to mitigate floods after learning from historical observed data.

\FIDLAr\ consists of two deep learning components.
The \texttt{Flood Manager} predicts control schedules, while the \texttt{Flood Evaluator} validates the above output by predicting the water levels resulting from these schedules. 
\FIDLAr\ has the following characteristics: (a) \FIDLAr\ makes effective use of reliable forecasts of specific variables (e.g., precipitation and tidal information) for the near future \cite{shi2023explainable}. (b) The training of \FIDLAr\ is treated as an optimization problem where its output is optimized by minimizing a loss function with an eye toward balancing flood mitigation and water wastage. (c) During training, \FIDLAr\ uses backpropagation from the Evaluator to improve the Manager by evaluating the generated schedules of the Manager. (d) \FIDLAr\ outputs control schedules for the hydraulic structures in the river system (e.g., gates and pumps) so as to achieve effective flood mitigation. After pre-training with historical data, \FIDLAr\ makes rapid predictions, achieving real-time flood mitigation.


\begin{figure}[ht]
\centering
\includegraphics[width=\columnwidth]{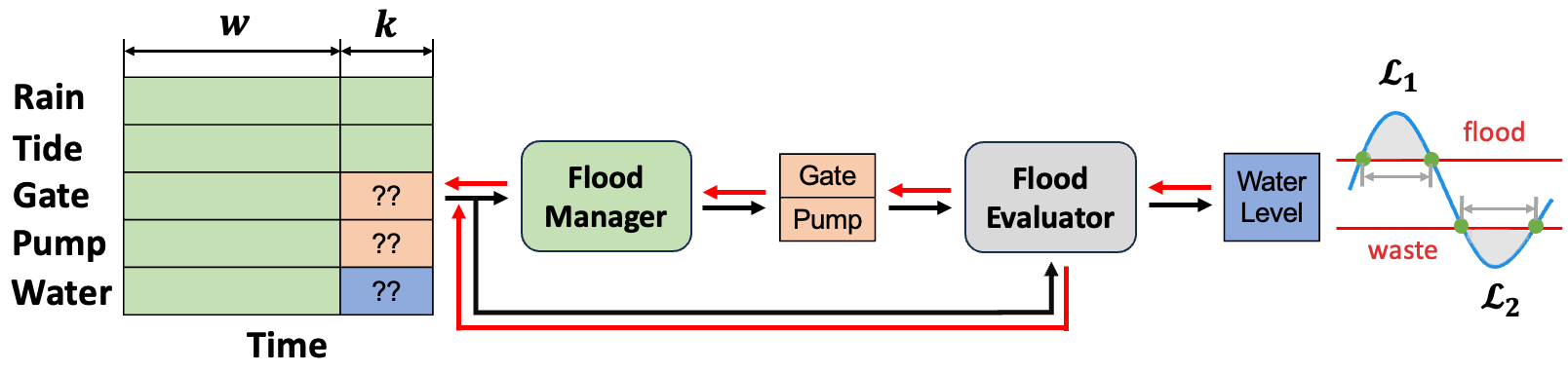} 
\caption{Forecast-Informed Deep Learning Architecture (\FIDLAr). Input data in the form of a table consisting of five variables as shown in the left end of the diagram. The variables $w$ and $k$ are the lengths of the past and prediction windows, respectively.
The parts of the table colored green are provided as inputs to the programs, while the orange and blue parts (with question marks) are outputs. The black and red arrows are the propagation and backpropagation processes. The \texttt{Flood Manager} and \texttt{Flood Evaluator} represent DL models, the former to predict the gate and pump schedules, and the latter to determine water levels for those schedules. Loss functions, $\mathcal{L}_{1}$ and $\mathcal{L}_{2}$, penalize the \emph{flooding} and \emph{water wastage} beyond thresholds, respectively. 
}
\label{fig:fidla_framework}
\end{figure}

\section{Problem Formulation}
Flood mitigation is achieved by predicting control schedules of hydraulic structures (gates and pumps) in the river system, $X^{gate, pump}_{t+1:t+k}$, for $k$ time points ($t+1$ through $t+k$) into the future, taking all the inputs, $X$, from the past $w$ time points, along with reliably forecasted covariates (rainfall and tide) for $k$ time points in the future. We train machine learning models to learn a function $f$ with parameters $\theta$ mapping the input variables to the control schedules. Thus,
\begin{equation}
\label{eq:problem_formulation}
    f_\theta: (X_{t-w+1:t}, X^{cov}_{t+1:t+k}) \rightarrow  X^{gate, pump}_{t+1:t+k},
\end{equation}
where the subscripts represent the time ranges under consideration, and the superscripts refer to the covariates in question. If the covariates are not mentioned, then all the variables are being considered. The superscript $cov$ refers to the covariates, rain and tides, both of which can be reliably predicted.

\section{Methodology}
We trained the ML model to learn the function, $f_{\theta}$, by treating this as an optimization problem. During training, the Flood Manager generates a sequence of control schedules for the gates and pumps, which the Flood Evaluator is used to predict the resulting future water levels, $X^{water}_{t+1:t+k}$ (see Eq. (\ref{eq:WaLeF})).
The \emph{backpropagation} algorithm \cite{lecun1988theoretical} is used to backpropagate the feedback on the quality of the generated schedules (using \emph{loss functions}, described in Section \ref{sec:loss}) to ``nudge'' the Flood Manager to produce more effective schedules. 
After the training of \FIDLAr\ is completed, the Flood Evaluator does not perform backpropagation but provides validation for the schedules.

\paragraph{Flood Evaluator}
The Flood Evaluator is used to predict water levels at specific locations of interest along the river systems (see Fig. \ref{fig:flood_evaluator}).
Its transfer function is described below.
\begin{equation}
\label{eq:WaLeF}
    g_\theta: (X_{t-w+1:t}, X^{cov}_{t+1:t+k}, X^{gate, pump}_{t+1:t+k}) \rightarrow X^{water}_{t+1:t+k}.
\end{equation}
The Flood Evaluator is pre-trained to predict water levels as accurately as possible for different conditions and control schedules. 
Note that the parameters of the pre-trained Flood Evaluator are immutable either during the training or testing operation of \FIDLAr.
It merely plays the role of a trained ``referee'' to evaluate those generated control schedules in \FIDLAr.
\paragraph{Flood Manager}
The Flood Manager is a DL-based model used to produce control schedules for hydraulic structures, taking reliably predictable future information (rain, tide) and all measured information from the recent past, as shown in Fig. \ref{fig:flood_manager}.
During training, the model parameters of Flood Manager are trained and optimized using the gradient descent algorithm with backpropagation from the Flood Evaluator \cite{ruder2016overview}.
The gradients are computed as partial derivatives of the loss functions (see Eq. (\ref{eq:flood_loss}) with respect to the input, i.e., green parts in Fig. \ref{fig:flood_manager}).

\paragraph{Loss function}
\label{sec:loss}
Loss functions are critical in steering the learning process to address the optimization goals. 
The metric of performance for the flood manager is the total time for which the water levels either exceed \textit{flooding threshold} or dip below \textit{water wastage threshold}.
Another related metric is the extent to which the limits are exceeded to signify the severity of floods or water wastage. 
\begin{equation}
\begin{aligned}
\label{eq:flood_loss}
    \mathcal{L}_{1} = \sum^{N}_{i=1} \sum^{t+k}_{j=t+1} \Arrowvert max\{X^{water}_{i, j} - X^{flood}_{i}, 0\} \Arrowvert^{2}, \\ 
    \mathcal{L}_{2} = \sum^{N}_{i=1} \sum^{t+k}_{j=t+1} \Arrowvert min\{X^{water}_{i, j} - X^{waste}_{i}, 0\} \Arrowvert^{2},
\end{aligned}
\end{equation}
where $N$ is the number of water level locations of interest; $k$ is the length of prediction horizon; $X^{flood}$ and $X^{waste}$ represent the thresholds for flooding and water wastage.  
The final loss function is a weighted combination of $\mathcal{L}_1$ and $\mathcal{L}_2$. 

\section{Experiments \& Results}
\paragraph{Flood Prediction} 
We compared eight DL models and one physics-based model (HEC-RAS) for the Flood Evaluator by predicting the water levels for a $k=24$-hour horizon. See Table \ref{tab:flood_prediction_s1} for the results.

\paragraph{Flood Mitigation} We tested eight DL models, two methods based on genetic algorithms (GA), and a rule-based baseline method, for flood mitigation.
See Table \ref{tab:flood_mitigate_s1} for the results. 


\section{Model Explainability}
\begin{figure}[ht]
\centering
\includegraphics[width=0.75\columnwidth]{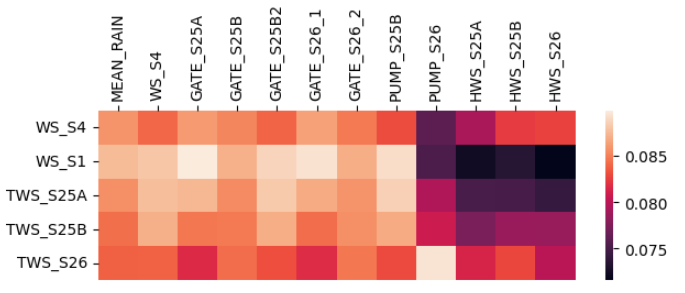} 
\caption{Attention scores between each variable of input. Labels such as \texttt{S4, S1, S25A, S25B, S26} represent locations shown on the map of the S. Florida watershed in Fig. \ref{fig:domain}. \texttt{WS\_S4} represents the water level measured at location \texttt{S4}, which also corresponds to the tidal information.}
\label{fig:water_covariate_attention}
\end{figure}

To investigate the relationship between covariates and water levels of input variables, Fig. \ref{fig:water_covariate_attention} visualizes the attention scores from the \texttt{Attention} layer \cite{vaswani2017attention} of our FloodGTN model in Fig. \ref{fig:graphtransformer}. It reveals all water levels at key locations rely mainly on the tides (\texttt{WS\_S4}). This makes sense when there are no structures between the location and the ocean. Besides, the water level at each location is also impacted by the structures close to it. These observations are consistent with prior knowledge.

\begin{figure}[h]
     \centering
     \begin{subfigure}[b]{\textwidth}
        \centering
        \includegraphics[scale=0.24]{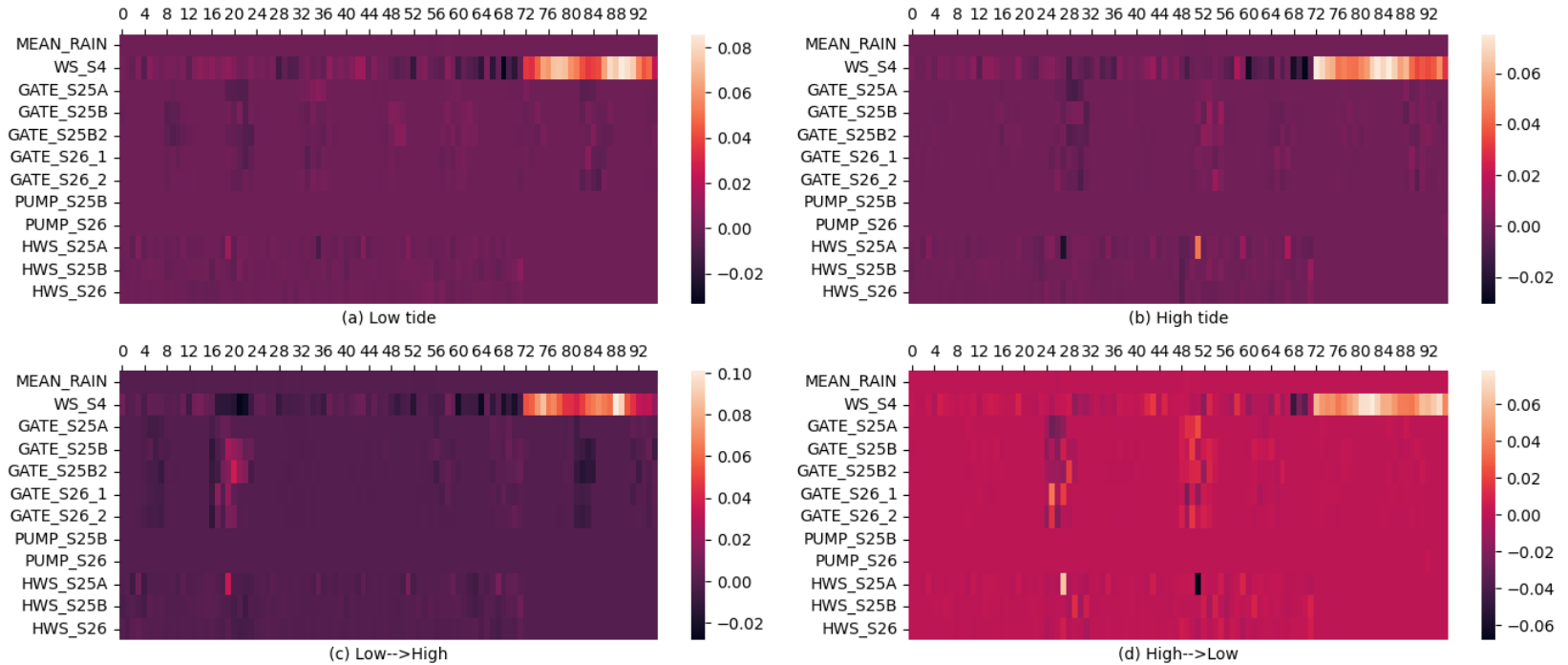}
        \caption{Contribution of each covariate to predict water levels for the next 24 hours.}
        \label{fig:covariate_contribution}
     \end{subfigure}
     \hfill
     \begin{subfigure}[b]{\textwidth}
        \centering
        \includegraphics[scale=0.215]{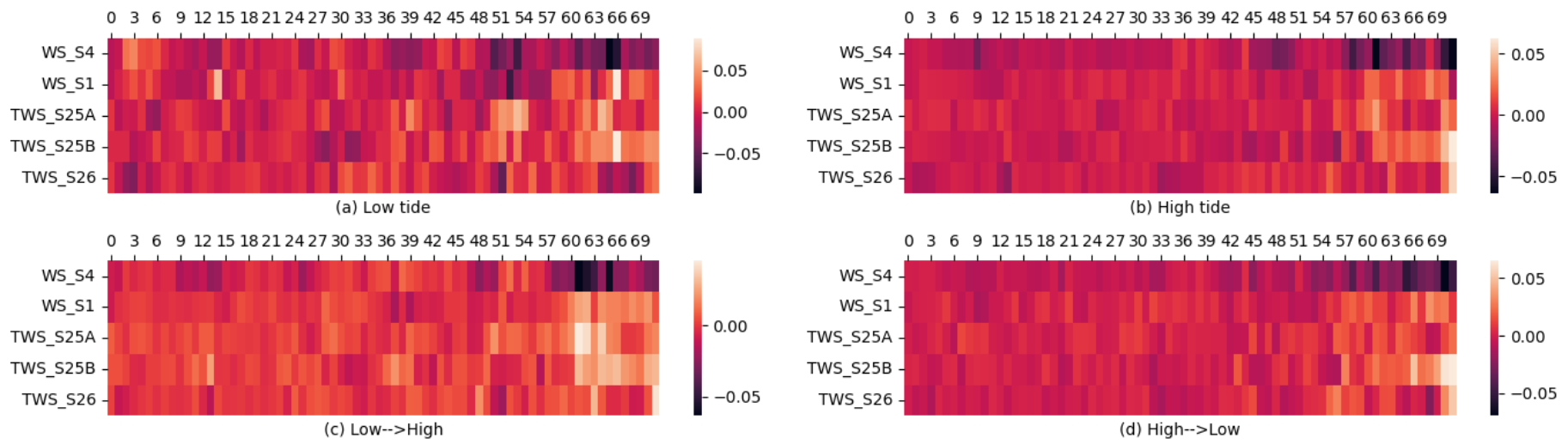}
        \caption{Contribution of water level readings for the past 3 days to predict water levels for the next 24 hours.}
        \label{fig:water_contribution}
     \end{subfigure}
        \caption{Model explainability for the contribution of input to output. Time point 72 is \texttt{NOW}, which ends based on the water levels at (a) low tide, (b) high tide, (c) low to high, and (d) high to low.}
        \label{fig:covariate_water_contribution}
\end{figure}

The contribution of each input variable (at each time point) on the model output was computed using \texttt{LIME} \cite{ribeiro2016should}. 
Fig. \ref{fig:covariate_water_contribution} shows these contributions as heatmaps.
First, what jumps out immediately is that the predicted water levels throughout the river system under normal conditions (mild to no rain) are overwhelmingly impacted by the tidal conditions measured at \texttt{S4}, which tend to be periodic.
The impact of the nearby hydraulic structures is next in significance. 
Second, the highest contributions come from the covariates from the last 24 columns, which correspond to the future predicted tidal data, showing categorically that the future estimates for the 24-hour horizon are invaluable.
The third critical insight is that FloodGTN focuses on the schedules of the gates, but only during low tide.
This makes sense because water pre-releases must happen during low tides since releasing water is typically not possible during high tides.
This was also confirmed by looking at the historical data (see Fig. \ref{fig:s4_gate25b}). 
Finally, we observe that predicted water levels also depend on data from the near past (see Fig. \ref{fig:water_contribution}), and this dependence wanes as we consider variables from the more distant past.


\vspace{-1.5mm}
\section{Conclusions}
\vspace{-1.5mm}
\label{sec:concl}
\FIDLAr\ is a DL-based tool to compute water ``pre-release'' schedules for hydraulic structures to achieve effective and efficient flood mitigation, while ensuring water wastage is avoided by managing the extent of pre-release. 
\FIDLAr\ consists of two DL-based components (\texttt{Flood Manager} \& \texttt{Flood Evaluator}). 
During training, backpropagation from the Evaluator helps or even forces the Manager to generate better outputs.
%
Finally, we summarize that all the DL-based versions of 
\FIDLAr\ are orders of magnitude faster than the (physics-based or GA-based) competitors, while achieving better flood mitigation. 
The use of explainability tools provides rare insights into a coastal system while validating DL models are learning correct and useful knowledge from the input.

\begin{ack}
This work is part of the I-GUIDE project, which is funded by the National Science Foundation under award number 2118329.
\end{ack}


\bibliographystyle{plain}
\bibliography{reference}

\appendix

\newpage

\section{Dataset}
\begin{figure}[ht]
\centering
\includegraphics[width=0.8\columnwidth]{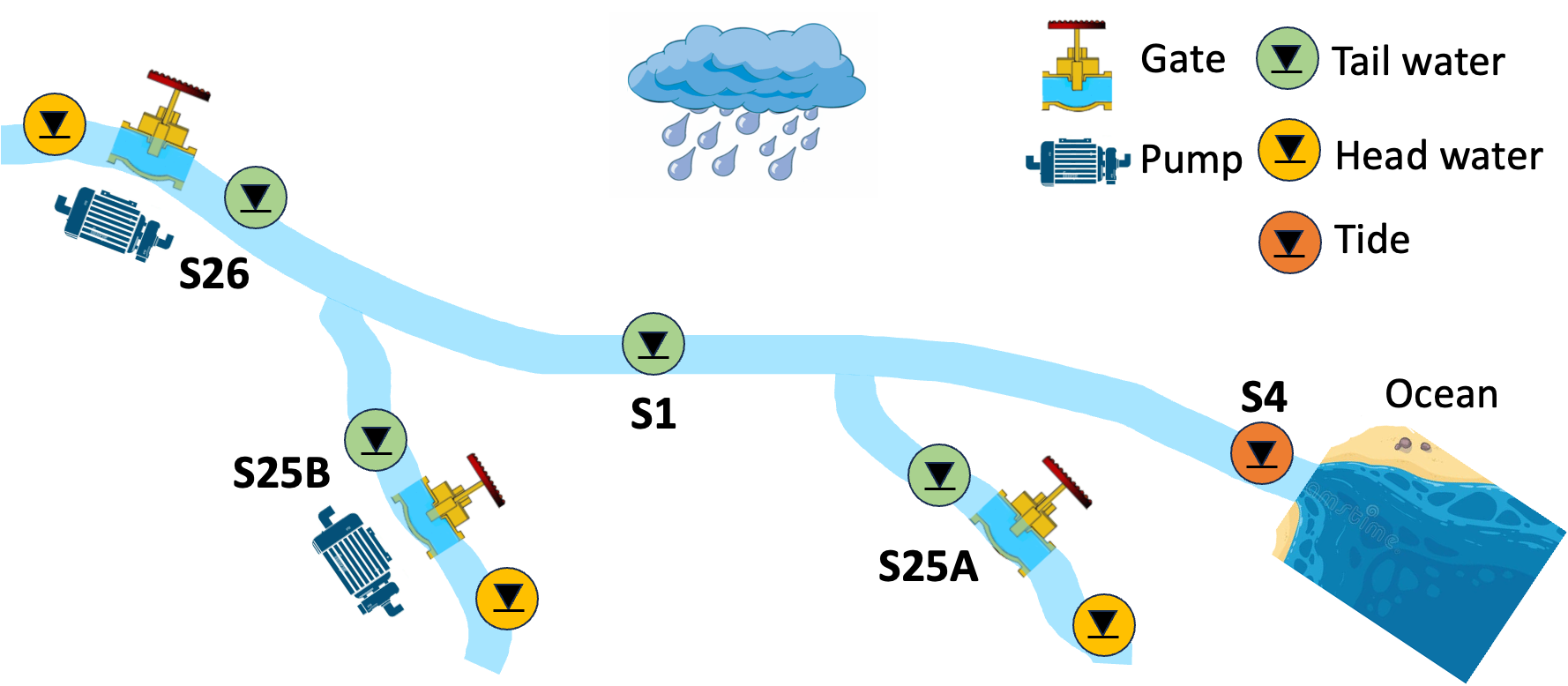} 
\caption{Schemetic diagram of the study domain - South Florida. There are three water stations with hydraulic structures (gates and pumps), one water station, and four control points of interest (labeled \texttt{S1, S25A, S25B, S26} and marked with a green circle), and one station (S4) monitoring tidal information.}
\label{fig:domain}
\end{figure}

We obtained data on a coastal section of the South Florida watershed from the South Florida Water Management District's (SFWMD) DBHydro database.
The data set we used in the work recorded the hourly observations for water levels and external covariates from 2010 to 2020.
As shown in Figure \ref{fig:domain}, the river system has three branches/tributaries and includes several hydraulic structures -- gates and pumps -- located along the river system to control water flows.
Water levels are also impacted by ocean tides since the river system empties itself into the ocean.
In this work, we aim to predict the effective schedules of the gates and pumps to mitigate or avoid flooding at four specific locations marked by green circles in Fig. \ref{fig:domain}.
It is useful to note that this portion of the river system flows through the large metropolis of Miami, which has a sizable population, commercial enterprises, and an international airport in its close vicinity.

\section{Model details}

\subsection{Framework of Flood Evaluator and Flood Manager} 
The architecture of the Flood Evaluator and Flood Manager are described here. The Flood Evaluator is used to predict the water levels based on the input of past information of all variables and any future covariates that may be estimated in advance. The future tide and rain information could be reliably predicted, while gate and pump information are decided by the operators. After pre-training, the Flood Evaluator can play the role of "scorer" to evaluate the quality of gate and pump schedules. The Flood Manager is used to generate the gate and pump schedules given the input of past information of all variables and estimated future covariates (i.e., rain and tide information). Variables $w$ and $k$ represent the length of the past and future horizons, respectively.

\begin{figure}[ht]
\centering
\includegraphics[width=0.57\columnwidth]{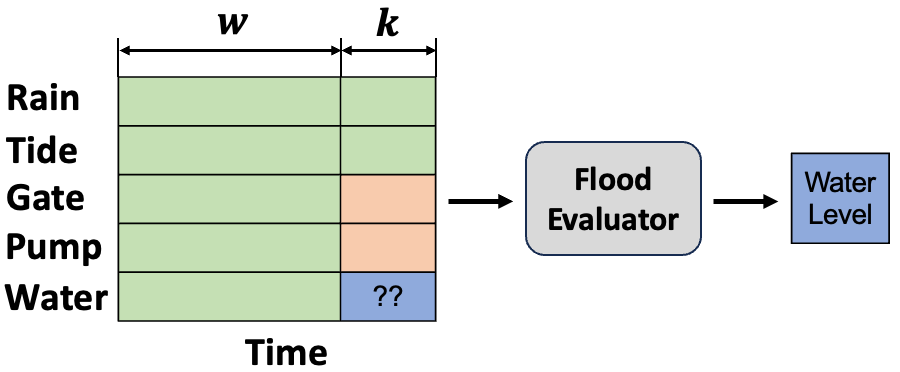} 
\caption{Flood Evaluator.}
\label{fig:flood_evaluator}
\end{figure}

\begin{figure}[ht]
\centering
\includegraphics[width=0.6\columnwidth]{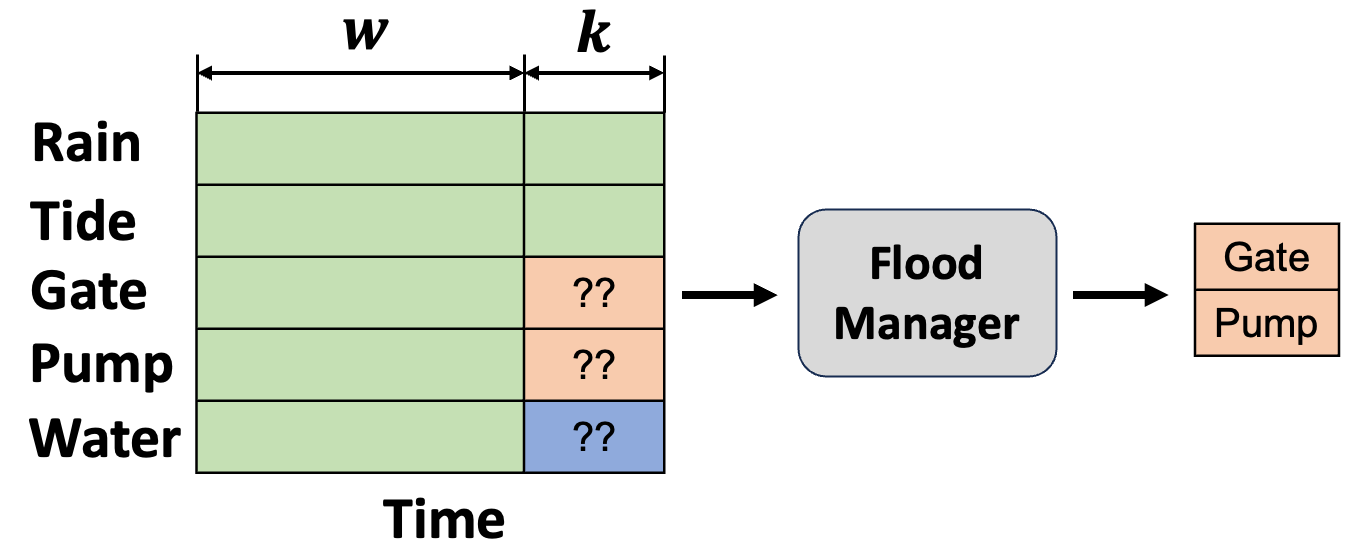} 
\caption{Flood Manager.}
\label{fig:flood_manager}
\end{figure}

\subsection{FloodGTN} 
The best-performing DL model for \underline{Flood} prediction and mitigation is described here and is referred to as FloodGTN (\underline{G}raph \underline{T}ransformer \underline{N}etwork). 
More specifically, FloodGTN combines graph neural networks (GNNs), attention-based transformer networks, long short-term memory networks (LSTMs), and convolutional neural networks (CNNs) for various objectives.
The GNN-LSTM model learns the spatio-temporal dynamics of water levels, while the CNN-Transformer module learns feature representations from the inputs.
Attention is used to figure out the interactions between covariates and water levels.
This model works best for both the flood evaluator and the manager.
\begin{figure}[ht]
\centering
\includegraphics[width=0.6\columnwidth]{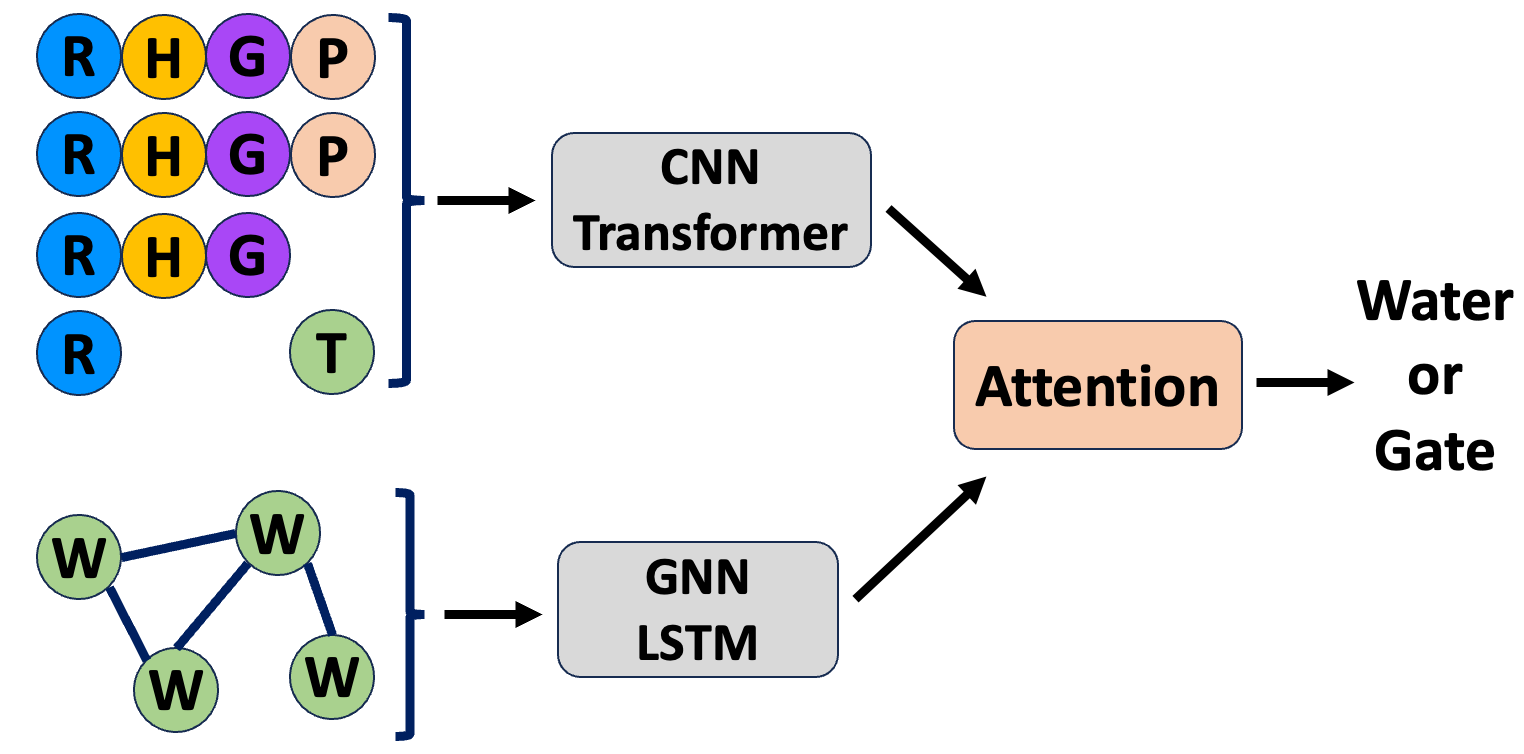} 
\caption{The Flood Graph Transformer Network (FloodGTN) model is assumed to have a collection of input variables, generically denoted by $R, T, G, P, H$, and $W$, measured at different measuring stations along the river system. The output variable is the set of water levels (for the Evaluator) or gate schedules (for the Manager). 
}
\label{fig:graphtransformer}
\end{figure}

\newpage

\section{Experimental results}
\subsection{Results for flood prediction}
The table below compares the performance of a graph-transformer-based evaluator tool (labeled FloodGTN) with the ground truth (measured data), physics-based HEC-RAS model, and seven other DL models.
\begin{table*}[ht!]
\centering
    \begin{tabular}{ccccccc}
    \toprule
    \textbf{Methods}    & \textbf{MAE (ft)}   & \textbf{RMSE (ft)} & \textbf{OverTime}   & \textbf{OverArea}  & \textbf{UnderTime}   & \textbf{UnderArea} \\
    \midrule\midrule
    Ground-truth      & -            & -        & 96       & 14.82     & 1,346    & 385.8  \\ 
    \midrule
    HEC-RAS           & 0.174        & 0.222    & 68       & 10.07    & 1,133    & 325.33  \\ 
    \midrule
    MLP               & 0.065        & 0.086    & 147      & 27.96    & 1,677    & 500.41  \\  
    RNN               & 0.054        & 0.072    & 110      & 17.12    & 1,527    & 441.41  \\   
    CNN               & 0.079        & 0.104    & 58       & 5.91     & 1,491    & 413.22  \\ 
    GNN               & 0.054        & 0.070    & 102      & 15.90    & 1,569    & 462.63  \\ 
    TCN               & 0.050        & 0.065    & 47       & 5.14     & 1,607    & 453.63  \\  
    RCNN              & 0.092        & 0.110    & 37       & 4.61     & 1,829    & 553.20  \\
    Transformer       & 0.050        & 0.066    & 151      & 25.95    & 1,513    & 434.13  \\ 
    \midrule
    FloodGTN          & 0.040        & 0.056    & 100      & 15.64    & 1,764    & 549.28  \\  
    \bottomrule
    \end{tabular}
\caption{Comparison of different models for the Flood Evaluator on test set (at time t+1 for measuring station S1). \emph{OverTime} (\emph{UnderTime}) represents the number of time steps during which the water levels exceed the upper threshold (subceed the lower threshold). Similarly, \emph{OverArea} (\emph{UnderArea}) refers to the area between the water level curve and the upper threshold (lower threshold).} 
\label{tab:flood_prediction_s1}
\end{table*}

\subsection{Results for flood mitigation}
The table below compares the performance of \FIDLAr\ (a graph-transformer-based flood mitigation tool using FloodGTN as an evaluator) with the rule-based method, two GA-based tools, and seven other DL models.

\begin{table*}[ht]
\centering
\begin{tabular}{cccccc}

\toprule
    & \textbf{Methods}   & \textbf{OverTime}    & \textbf{OverArea}   & \textbf{UnderTime}    & \textbf{UnderArea}  \\
\midrule \midrule     
    & Rule-based             & 96         & 14.82     & 1,346     & 385.8  \\  
\midrule
\multirow{2}{*}{GA-Based}    & Genetic Algorithm$^{*}$     & -          & -         & -         & -  \\
    & Genetic Algorithm$^{\dag}$      & 86         & 16.54     & 454       & 104  \\
\midrule
\multirow{8}{*}{DL-Based}       & MLP                    & 91         & 13.31     & 1,071     & 268.35 \\  
    & RNN                    & 35         & 3.97      & 251       & 41.05  \\   
    & CNN                    & 81         & 11.22     & 1,163     & 314.37  \\ 
    & GNN                    & 31         & 3.72      & 429       & 84.31  \\ 
    & TCN                    & 39         & 3.77      & 306       & 55.12  \\  
    & RCNN                   & 29         & 3.28      & 328       & 58.68  \\
    & Transformer            & 85         & 11.54     & 1,180     & 310.16  \\ 
    & \FIDLAr/FloodGTN               & 22         & 2.23      & 299       & 53.34  \\
\bottomrule
\end{tabular}
\caption{Comparison of the Flood Manager tool, \FIDLAr\ on test set (at time t+1 for station S1). The GA method with a $*$ was used with a physics-based (HEC-RAS) as the Evaluator, while the GA method with a $\dag$ was used with the DL-based FloodGTN as the Evaluator. All other rows are DL-based flood managers with FloodGTN as the Evaluator.}
\label{tab:flood_mitigate_s1}
\end{table*}

\pagebreak
\subsection{Results for flood mitigation for a small event at a certain location}

\begin{figure}[ht]
\centering
\includegraphics[width=0.53\columnwidth]{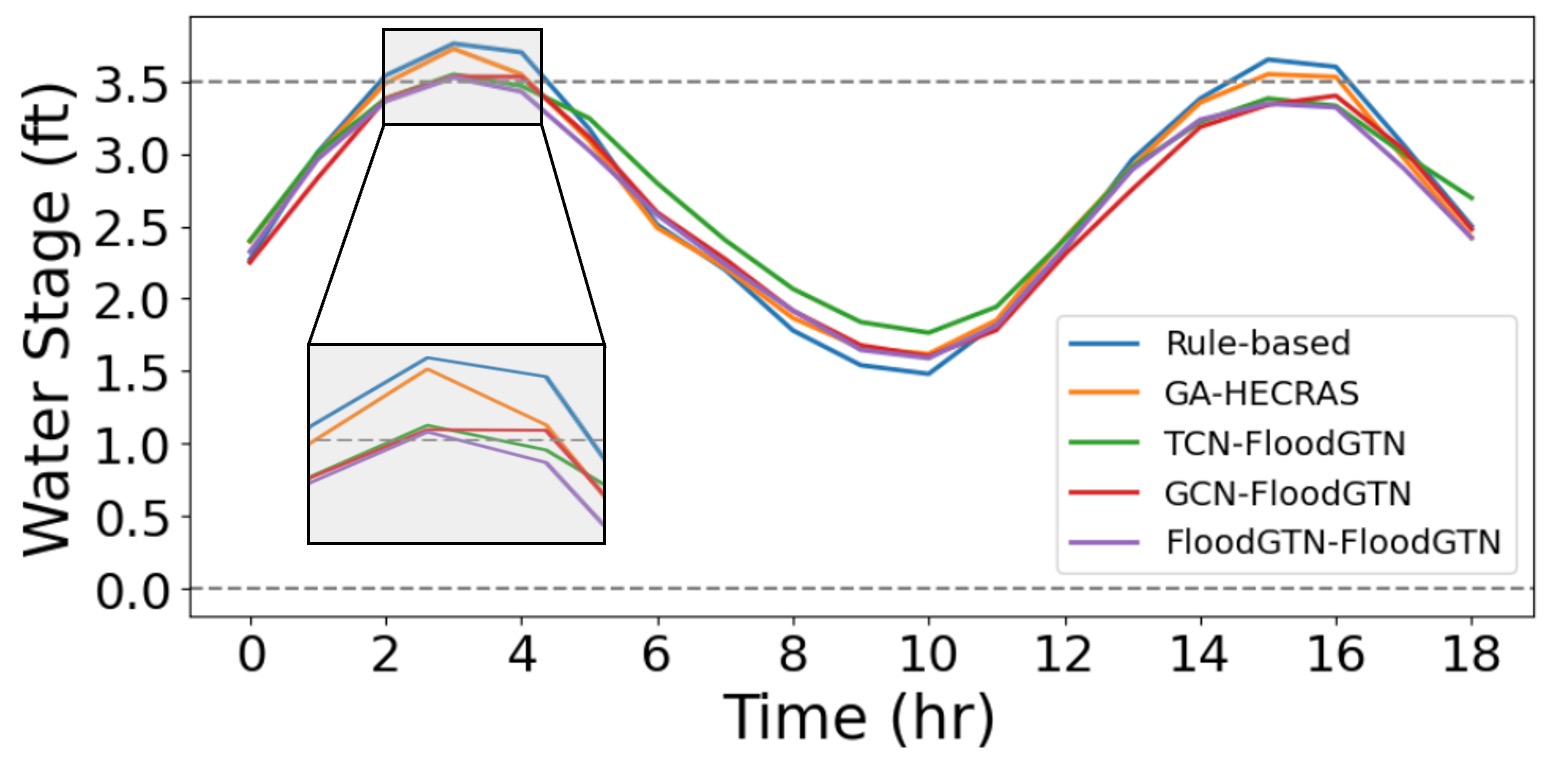} 
\caption{Flood mitigation visualization for a short sample spanning 18 hours from Sep. 3rd (09:00) to Sep. 4th (03:00) in 2019 for one location of interest. We zoomed in on the time period $t=2$ to 4.5 hours, shaded in gray.}
\label{fig:visualize_flood_mitigation_s1}
\end{figure}

Here we visualize the water levels with the rule-based and the DL-based approaches \FIDLAr\ for a short sample for one location of interest in Figure \ref{fig:visualize_flood_mitigation_s1}. 
The zoomed portion shows a 2.5-hour period where the floods are mitigated to bring water levels to under 3.5 feet based on the predicted gate and pump schedules.
The corresponding performance measures for this small sample are provided in Table \ref{tab:flood_mitigate_event_s1}.

Table \ref{tab:flood_mitigate_event_s1} below shows the corresponding results for Fig. \ref{fig:visualize_flood_mitigation_s1}.
\begin{table*}[ht]
\centering
    \begin{tabular}{cccc}
    \toprule
                                & \textbf{Methods}   & \textbf{Over Timesteps}    & \textbf{Over Area}  \\
    \midrule \midrule     
    \multirow{1}{*}{}           & Rule-based     & 6      & 0.866      \\
    \midrule
    \multirow{2}{*}{GA-Based}   & Genetic Algorithm$^{*}$            & 4      & 0.351      \\
                                & Genetic Algorithm$^{\dag}$        & 6      & 0.764      \\
    \midrule
    \multirow{8}{*}{DL-Based}   & MLP          & 6      & 0.614      \\  
                                & RNN          & 1      & 0.074      \\   
                                & CNN          & 6      & 0.592      \\ 
                                & GNN          & 2      & 0.062      \\ 
                                & TCN          & 1      & 0.046      \\  
                                & RCN         & 1      & 0.045      \\
                                & Transformer  & 6      & 0.614      \\ 
                                & \FIDLAr/FloodGTN     & 1      & 0.022      \\
    \bottomrule
    \end{tabular}
\caption{Comparison of different methods of Flood Manager for flood Mitigation (at time t+1 for one event S1). The experimental results correspond to the time period displayed in Fig. \ref{fig:visualize_flood_mitigation_s1}.
The GA method with a $*$ was used with a physics-based (HEC-RAS) as the Evaluator, while the GA method with a $\dag$ was used with the DL-based FloodGTN as the Evaluator. All other rows are DL-based flood managers with FloodGTN as the Evaluator.}
\label{tab:flood_mitigate_event_s1}
\end{table*}

\newpage
\subsection{Computational time}
Table \ref{tab:flood_prediction_time} shows the running times for all the methods for the evaluator component and for the whole flood mitigation system in its test phase. 
All the DL-based approaches are several orders of magnitude faster than the currently used physics-based and GA-based approaches. 
The table also shows the training times for the DL-based approaches, which do not impact the real-time performance, once deployed.
\begin{table*}[ht]
\centering
  \begin{tabular}{ccccc}
    \toprule
    \multirow{2}{*}{\textbf{Model}}  & \multicolumn{2}{c}{\textbf{Prediction}}        & \multicolumn{2}{c}{\textbf{Mitigation}} \\
                       & \textbf{Train}     & \textbf{Test}  & \textbf{Train}   & \textbf{Test} \\
    \midrule\midrule
    
    HEC-RAS            & -              & 45 min      & -              & -   \\ 
    Rule-based         & -              & -           & -              & -   \\
    GA$^{*}$           & -              & -           & -              & -     \\ 
    GA$^{\dag}$        & -              & -           & -              & est. 30 h  \\ 
    \midrule
    MLP                & 35 min         & 1.88 s      & 58 min         & 6.13 s \\
    RNN                & 243 min        & 8.57 s      & 54 min         & 12.75 s \\
    CNN                & 37 min         & 1.93 s      & 17 min         & 5.84 s \\
    GNN                & 64 min         & 3.13 s      & 29 min         & 7.26 s \\
    TCN                & 60 min         & 4.57 s      & 45 min         & 9.06 s \\
    RCNN               & 136 min        & 8.61 s      & 61 min         & 13.27 s \\
    Transformer        & 43 min         & 2.38 s      & 23 min         & 6.76 s \\
    FloodGTN           & 119 min        & 2.95 s      & 35 min         & 4.90 s \\
    \bottomrule
  \end{tabular}
\caption{The running times for flood prediction and mitigation. The running times for the rule-based method are not reported since historical data was directly used. GA$^{*}$, which combines a GA-based tool and HEC-RAS, took too long and was not reported. GA$^{\dag}$, which combines the GA-based tool with FloodGTN, also took too long but was estimated using a smaller sample.}
\label{tab:flood_prediction_time}
\end{table*}

\section{Visualization of observed variables}
We visualize the observed variables, \texttt{WS\_S4}, \texttt{Gate\_S25B}, for a better understanding of explainability in Fig. \ref{fig:covariate_contribution}.
\begin{figure}[ht]
\centering
\includegraphics[width=0.94\columnwidth]{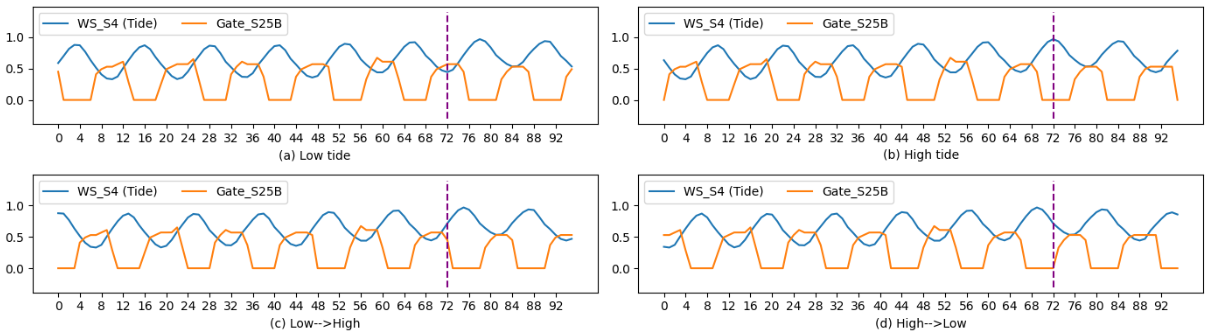} 
\caption{Visualization of input variables: \texttt{WS\_S4} (tide information) and \texttt{Gate\_S25B} (gate schedules at location S25B). Time point 72 is \texttt{NOW}.}
\label{fig:s4_gate25b}
\end{figure}

\end{document}